\documentclass[10pt,twocolumn,letterpaper]{article}

\usepackage{cvpr}              

\usepackage{graphicx}
\usepackage{amsmath}
\usepackage{amssymb}
\usepackage{booktabs}
\usepackage{multirow}
\usepackage{xcolor}
\definecolor{mygreen}{RGB}{83,161,81}
\definecolor{myred}{RGB}{178,34,34}
\usepackage[pagebackref,breaklinks,colorlinks]{hyperref}

\usepackage[capitalize]{cleveref}
\crefname{section}{Sec.}{Secs.}
\Crefname{section}{Section}{Sections}
\Crefname{table}{Table}{Tables}
\crefname{table}{Tab.}{Tabs.}

\def\confName{CVPR}
\def\confYear{2022}

\usepackage{overpic}
\usepackage{enumitem} 
\usepackage{overpic} 
\usepackage{color}

\definecolor{turquoise}{cmyk}{0.65,0,0.1,0.3}
\definecolor{purple}{rgb}{0.65,0,0.65}
\definecolor{dark_green}{rgb}{0, 0.5, 0}
\definecolor{orange}{rgb}{0.8, 0.6, 0.2}
\definecolor{red}{rgb}{0.8, 0.2, 0.2}
\definecolor{darkred}{rgb}{0.6, 0.1, 0.05}
\definecolor{blueish}{rgb}{0.0, 0.3, .6}
\definecolor{light_gray}{rgb}{0.7, 0.7, .7}
\definecolor{pink}{rgb}{1, 0, 1}
\definecolor{greyblue}{rgb}{0.25, 0.25, 1}

\usepackage{blindtext}

\renewcommand{\paragraph}[1]{\vspace{1em}\noindent\textbf{#1}.}
\begin{document}
\title{MetaFormer : A Unified Meta Framework for Fine-Grained Recognition}

\author{
Qishuai Diao$^1$,
~~~
Yi Jiang$^1$,
~~~
Bin Wen$^1$,
~~~
Jia Sun$^1$,
~~~
Zehuan Yuan$^1$\\
~~~
${^1}$ByteDance Inc. ~~~
}
\maketitle
\begin{abstract}
Fine-Grained Visual Classification (FGVC) is the task that requires recognizing the objects belonging to multiple subordinate categories of a super-category. Recent state-of-the-art methods usually design sophisticated learning pipelines to tackle this task. However, visual information alone is often not sufficient to accurately differentiate between fine-grained visual categories. Nowadays, the meta-information (e.g., spatio-temporal prior, attribute, and text description) usually appears along with the images. This inspires us to ask the question: Is it possible to use a unified and simple framework to utilize various meta-information to assist in fine-grained identification? To answer this problem, we explore a unified and strong meta-framework (\textbf{MetaFormer}) for fine-grained visual classification. In practice, MetaFormer provides a simple yet effective approach to address the joint learning of vision and various meta-information. Moreover, MetaFormer also provides a strong baseline for FGVC without bells and whistles.
Extensive experiments demonstrate that MetaFormer can effectively use various meta-information to improve the performance of fine-grained recognition. 
In a fair comparison, MetaFormer can outperform the current SotA approaches with only vision information on the iNaturalist2017 and iNaturalist2018 datasets. Adding meta-information, MetaFormer can exceed the current SotA approaches by \textbf{5.9\%} and \textbf{5.3\%}, respectively. Moreover, MetaFormer can achieve 92.3\% and 92.7\% on CUB-200-2011 and NABirds, which significantly outperforms the SotA approaches. The source code and pre-trained models are released at \url{https://github.com/dqshuai/MetaFormer}.

\end{abstract}



\section{Introduction}
In contrast to generic object classification, fine-grained visual classification aims to correctly classify objects belonging to the same basic category (birds, cars, etc.) into subcategories. FGVC has long been considered a challenging task due to the small inter-class variations and large intra-class variations. 

To the best of our knowledge, predominant approaches for FGVC are mainly concerned about how to make the network focus on the most discriminative regions, such as part-based model \cite{ge2019weakly, liu2020filtration,ding2019selective} and attention-based model \cite{fu2017look,zheng2017learning}. Intuitively, such methods introduce inductive bias of localization to neural networks with elaborate structure, inspired by human observation behavior. In addition, human experts often use information besides vision to assist them in classifying when some species are visually indistinguishable. Note that the data of fine-grained recognition is multi-source heterogeneous in the era of information explosion. Therefore, it is unreasonable that the neural network completes fine-grained classification tasks only with visual information. In practice, fine-grained classification, which is more difficult to distinguish visually, requires the help of orthogonal signals more than coarse-grained classification. Previous work \cite{chu2019geo,mac2019presence,he2017fine} utilize additional information, such as spatio-temporal prior and text description, to assist fine-grained classification. However, the design of these works for additional information only targets specific information, which is not universal. This inspires us to design a unified yet effective method to utilize various meta-information flexibly. 

\begin{figure}[htbp]
    \begin{center}
        \includegraphics[scale=0.6]{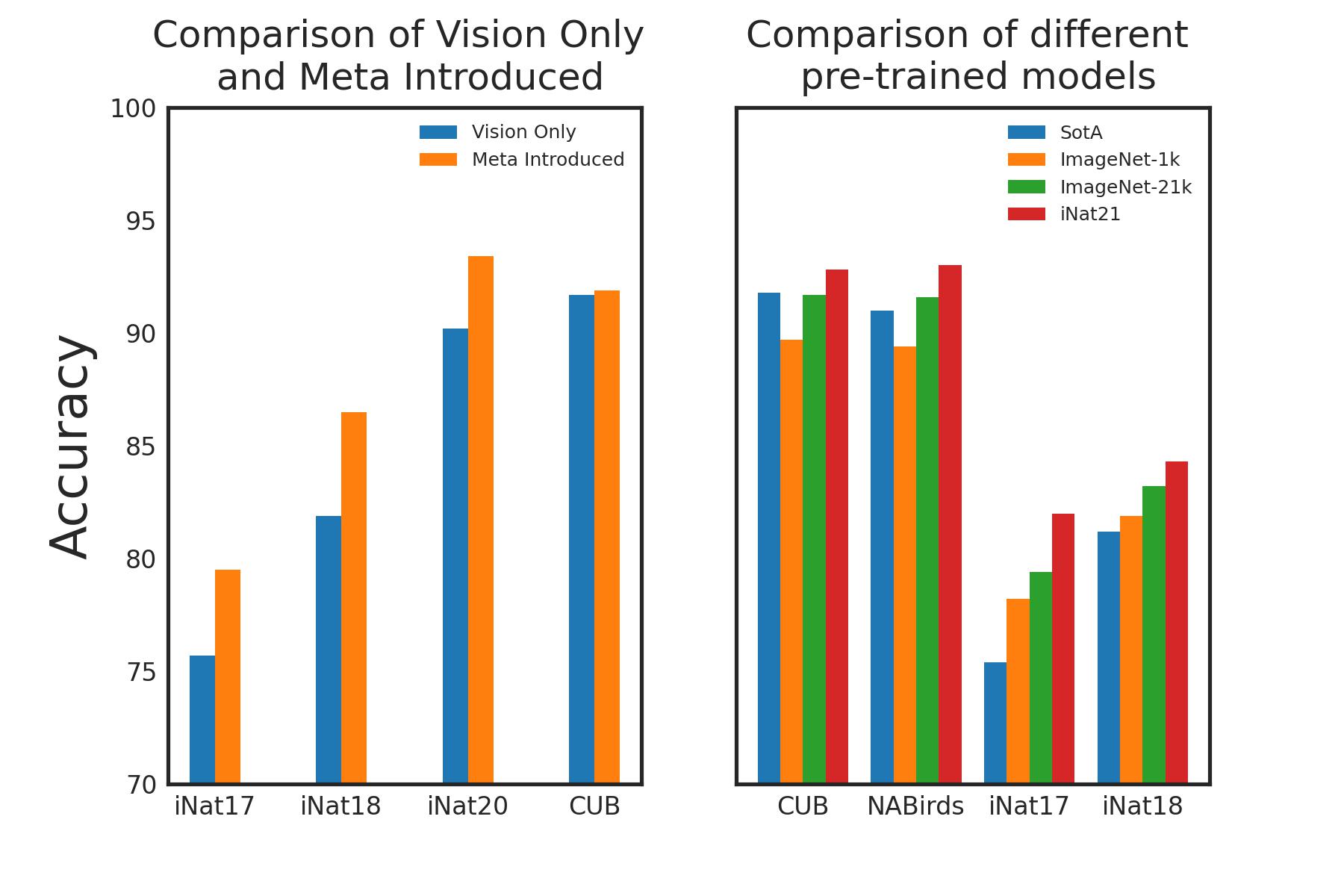}
    \end{center}
\caption{An overview of performance comparison of MetaFormer which using various meta-information and large-scale pre-trained model with state-of-the-art methods.}
\label{performance}
\end{figure}

Vision Transformer (ViT) shows pure transformer applied directly to sequences of image patches can perform very well on image classification tasks. Intuitively, it is feasible to simultaneously take vision token and meta token as the input of the transformer for FGVC. However, it is still unclear whether the different modalities impair the model's performance when interfering with each other. To answer this problem, we propose MetaFormer which uses a transformer to fuse vision and meta-information. As shown in Figure \ref{performance}, MetaFormer can effectively improve the accuracy of FGVC with the assistance of meta-information. In practice, MetaFormer can also be seen as a hybrid structure backbone where the convolution can downsample the image and introduce the inductive bias of the convolution, and the transformer can fuse visual and meta-information. In this manner, MetaFormer also provides a strong baseline for FGVC without bells and whistles.

Recent advances in image classification \cite{he2021transfg, ridnik2021imagenet} demonstrate large-scale pre-training could effectively improve the accuracy of both coarse-grained classification and fine-grained classification. However, most of the current methods for FGVC are based on ImageNet-1k for pre-training, which hinders further exploration of fine-grained recognition. 
Thanks to the simplicity of MetaFormer, we further explore the influence of the pre-trained model in detail, which can provide references to researchers regarding the pre-trained model. As shown in Figure \ref{performance}, large-scale pre-trained models can significantly improve the accuracy of fine-grained recognition. Surprisingly, without introducing any priors for fine-grained tasks, MetaFormer can achieve the SotA performance on multiple datasets using the large-scale pre-trained model.

The contribution of this study are summarized as follows:
\begin{itemize}
    \item We propose a unified and extremely effective meta-framework for FGVC to unify the visual appearance and various meta-information. This urges us to reflect on the development of FGVC from a brand fresh perspective.
    \item We provide a strong baseline for FGVC by only using the global feature. Meanwhile, we explored the impact of the pre-trained model on fine-grained classification in detail. Code and pre-trained models are available to assist researchers in further exploration.
    \item Without any inductive bias of fine-grained visual classification task, MetaFormer can achieve 92.3\% and 92.7\% on CUB-200-2011 and NABirds, outperforming the SotA approaches. Using only vision information, MetaFormer can also achieve SotA performance (78.2\% and 81.9\%) on iNaturalist 2017 and iNaturalist 2018 in a fair comparison. 
\end{itemize}






\section{Related Work}
In this section, we briefly review existing works on fine-grained visual classification and transformer.
\subsection{Fine-Grained Visual classification}
The existing fine-grained classification methods can be divided into vision only and multi-modality. The former relies entirely on visual information to tackle the problem of fine-grained classification, while the latter tries to take multi-modality data to establish joint representations for incorporating multi-modality information, facilitating fine-grained recognition.

\textbf{Vision Only.} Fine-grained classification methods that only rely on vision can be roughly classified into two categories: localization methods \cite{ ge2019weakly,liu2020filtration,zheng2019looking} and feature-encoding methods \cite{yu2018hierarchical,zheng2019learning,zhuang2020learning}. Early work \cite{luo2019cross, wei2018mask} used part annotations as supervision to make the network pay attention to the subtle discrepancy between some species and suffers from its expensive annotations. RA-CNN \cite{fu2017look} was proposed to zoom in subtle regions, which recursively learns discriminative region attention and region-based feature representation at multiple scales in a mutually reinforced way. MA-CNN \cite{zheng2017learning} designed a multi-attention module where part generation and feature learning can reinforce each other. 
NTSNet \cite{yang2018learning} proposed a self-supervision mechanism to localize informative regions without part annotations effectively.
Feature-encoding methods are devoted to enriching feature expression capabilities to improve the performance of fine-grained classification. Bilinear CNN \cite{lin2015bilinear} was proposed to extract higher-order features, where two feature maps are multiplied using the outer product. 
HBP \cite{yu2018hierarchical} further designed a hierarchical framework to do cross-layer bilinear pooling. DBTNet \cite{zheng2019learning} proposed deep bilinear transformation, which takes advantage of semantic information and can obtain bilinear features efficiently.
CAP \cite{behera2021context} designed context-aware attentional pooling to captures subtle changes in image. TransFG \cite{he2021transfg} proposed a Part Selection Module to select discriminative image patches applying vision transformer.
Compared with localization methods, feature-encoding methods are difficult to tell us the discriminative regions between different species explicitly.

\textbf{Multi Modality.} In order to differentiate between these challenging visual categories, it is helpful to take advantage of additional information, i.e., geolocation, attributes, and text description. Geo-Aware \cite{chu2019geo} introduced geographic information prior to fine-grained classification and systematically examined a variety of methods using geographic information prior, including post-processing, whitelisting, and feature modulation. Presence-Only \cite{mac2019presence} also introduced spatio-temporal prior into the network, proving that it can effectively improve the final classification performance. KERL \cite{chen2018knowledge} combined rich additional information and deep neural network architecture, which organized rich visual concepts in the form of a knowledge graph. Meanwhile, KERL \cite{chen2018knowledge} used a gated graph neural network to propagate node messages through the graph to generate knowledge representation. CVL \cite{he2017fine} proposed a two-branch network where one branch learns visual features, one branch learns text features, and finally combines the two parts to obtain the final latent semantic representations. The methods mentioned above are all designed for specific prior information and cannot flexibly adapt to different auxiliary information.
\subsection{Vision Transformer}
Transformer was first proposed for machine translation by \cite{vaswani2017attention} and has since been become a general method in natural language processing. Inspired by this, transformer models are further extended to other popular computer vision tasks such as object detection \cite{carion2020end, sparsercnn}, segmentation \cite{setr}, object tracking\cite{sun2020transtrack, trackformer}, video instance segmentation\cite{vistr, seqformer}. Lately, Vision Transformer (ViT) \cite{dosovitskiy2020image} directly applied pure transformer to the image patch for classification and achieved impressive performance. Compared with CNN, Vision Transformer has much less image-specific inductive bias. As a result, ViT requires large-scale training datasets (i.e., JFT-300M), intense data augmentation, and regularization strategies to perform well. Following ViT, \cite{d2021convit,liu2021swin} tried to introduce some inductive bias, i.e., convolutional inductive biases, and locality into the vision transformer.


\section{Method}
\label{method}
We introduce the hybrid framework that combines convolution and vision transformer in section \ref{hybrid framework}. Then, section \ref{meta information} elaborates on how to add meta-information to improve the performance of fine-grained classification.
\subsection{Hybrid Framework}
\label{hybrid framework}

\begin{figure*}[htbp]
    \begin{center}
        \includegraphics[scale=0.3]{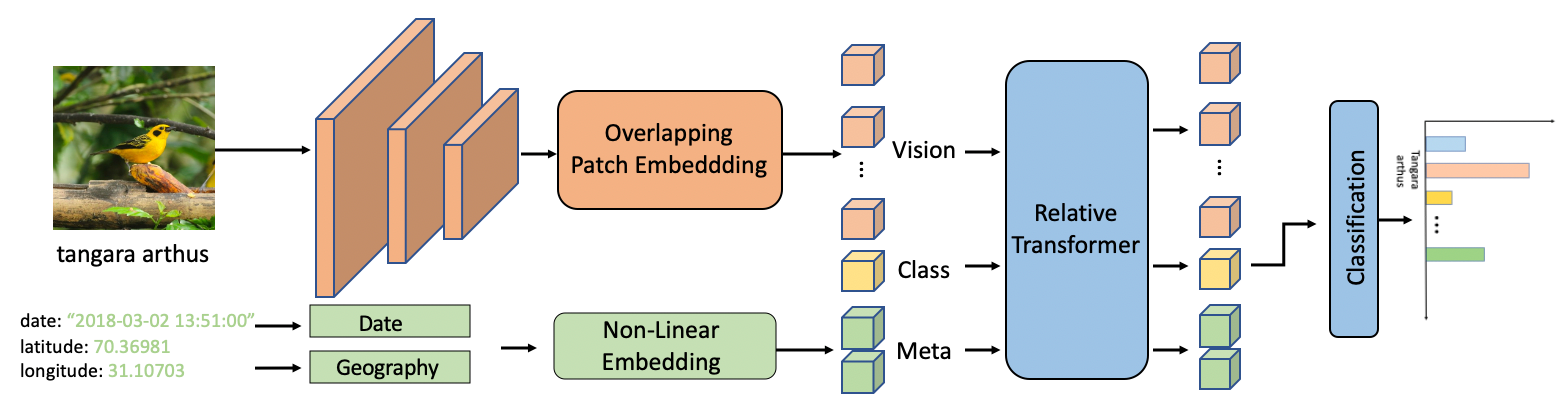}
    \end{center}
\caption{The overall framework of MetaFormer with meta-information. MetaFormer can also be seen as a pure backbone for FGVC except Non-Linear Embedding. The meta-information is encoded by non-linear embedding. Vision token, Meta token and Class token are used for information fusion through the Relative Transformer Layer. Finally, the class token is used for the category prediction.}
\label{fig:overview}
\end{figure*}

The overall framework of MetaFormer is shown in Fig \ref{fig:overview}. In practice, MetaFormer is a hybrid framework where convolution is used to encode vision information, and the transformer layer is used to fuse vision and meta information. Following canonical ConvNet, we construct a network of 5 stages (S0, S1, S2, S3\&S4). At the beginning of each stage, the input size decreases to realize the layout of different scales. The first stage S0 is a simple 3-layer convolutional stem. In addition, S1 and S2 are MBConv blocks with squeeze-excitation. We employ Transformer blocks with relative position bias in S3 and S4. Starting from S0 to S4, we always reduce the input size by 2$\times$ and increase the number of channels. The downsampling of s3 and s4 is convolution with stride 2, also known as Overlapping Patch Embedding. Following \cite{dai2021coatnet}, details of MetaFormer series as summarized in Table \ref{table:architecture}.


\begin{table}[htbp]
\caption{Detail setting of MetaFormer series. L denotes the number of blocks, and D represents the hidden dimension (channels).}
\label{table:architecture}
\setlength{\tabcolsep}{1mm}{
\begin{tabular}{c|cc|cc|cc}
\hline
\textbf{Stages} & \multicolumn{2}{c|}{\textbf{MetaFormer-0}} & \multicolumn{2}{c|}{\textbf{MetaFormer-1}} & \multicolumn{2}{c}{\textbf{MetaFormer-2}} \\ \hline
\textbf{S0}     & L=3            & D=64             & L=3             & D=64            & L=3            & D=128           \\
\textbf{S1}     & L=2            & D=96             & L=2             & D=96            & L=2            & D=128           \\
\textbf{S2}     & L=3            & D=192            & L=6             & D=192           & L=6            & D=256           \\
\textbf{S3}     & L=5            & D=384            & L=14            & D=384           & L=14           & D=512           \\
\textbf{S4}     & L=2            & D=768            & L=2             & D=768           & L=2            & D=1024          \\ \hline
\end{tabular}
}
\end{table}

\textbf{Relative Transformer Layer.} The self-attention operation in Transformer is permutation-invariant, which cannot leverage the order of the tokens in an input sequence. To mitigate this problem, following \cite{bao2020unilmv2,raffel2019exploring}, we introduce a relative position bias $B \in \mathbb{R}^{(M^2+N) \times (M^2+N)}$ to each position in computing similarity as follows:
\begin{equation}
    \label{equ:self-attention}
    Attention(Q,K,V) = SoftMax(QK^T/\sqrt{d}+B)V
\end{equation}
where $Q,K,V \in \mathbb{R}^{(M^2+N) \times d}$ are \textit{query}, \textit{key} and \textit{value} matrices. $M^2$ is the number of patches of the image. $N$ is the number of extra tokens, including class token and meta tokens. $d$ is the \textit{query/key} dimension. Follow \cite{liu2021swin}, we parameterize a matrix $\hat{B} \in \mathbb{R}^{(2M-1) \times (2M-1) + 1}$, since the relative position of the image block varies from $-M-1$ to $M+1$ and a special relative position bias is needed to indicate the relative position of the extra token and the vision token. There is no spatial position relationship between each extra token and other tokens, so all extra tokens only share the same relative position bias.
The relative transformer block (Eq. \ref{equ:transformer block}) contains multihead self-attention with relative position bias (MSA), multi-layer perceptron (MLP) blocks and Layernorm (LN). $\textbf{z}_0$ in Eq. \ref{equ:transformer block} represents the token sequence including classification token ($\textbf{x}_{class}$), meta token ($\textbf{x}^{i}_{meta}$) and visual token ($\textbf{x}^{i}_{vision}$).
\begin{equation}
    \label{equ:transformer block}
    \begin{split}
        &\textbf{z}_0 = [\textbf{x}_{class};\textbf{x}^{1}_{meta}, \cdots , \textbf{x}^{n-1}_{meta};\textbf{x}^{1}_{vision}, \cdots , \textbf{x}^{m}_{vision}] \\
        &\textbf{z}^{'}_i = MSA(LN(\textbf{z}_{i-1})) + \textbf{z}_i \\
        &\textbf{z}_i = MLP(LN(\textbf{z}^{'}_i)) + \textbf{z}^{'}_i  \quad \textbf{z}_i \in \mathbb{R}^{(M^2+N) \times d}
    \end{split}
\end{equation}

\textbf{Aggregate Layer.} S3 and S4 output two class tokens $\textbf{z}^1_{class}$ and $\textbf{z}^2_{class}$ at the end, respectively, which represent the fusion of vision features and meta-information. Note that the dimension of $\textbf{z}^1_{class}$ and $\textbf{z}^2_{class}$ are different, hence $\textbf{z}^1_{class}$ is expanded by MLP. Next, $\textbf{z}^1_{class}$ and $\textbf{z}^2_{class}$ are aggregated by Aggregate Layer which is as follows:
\begin{equation}
    \label{equ:aggregate layer}
    \begin{split}
        &\hat{\textbf{z}}^1_{class} = MLP(LN(\textbf{z}^1_{class}))\\
        &\textbf{z}_{class} = Conv1d(Concat(\hat{\textbf{z}}^1_{class},\textbf{z}^2_{class}))\\
        &\textbf{y} = LN(\textbf{z}_{class})
    \end{split}
\end{equation}
where $\textbf{y}$ is the output that combines multi-scale vision and meta information.

\textbf{Overlapping Patch Embedding.} We use overlapping patch embedding to tokenize the feature map and implement downsampling to reduce computational consumption. Following \cite{wang2021pvtv2}, we use convolution with zero padding to implement overlapping patch embedding as well. 

\subsection{Meta Information}
\label{meta information}
Relying on appearance information alone is often not sufficient to accurately distinguish some fine-grained species. When an image of species is given, human experts also make full use of additional information to assist in making the final decision. Recent advances in Vision Transformer show that it is feasible to encode images into sequence tokens in computer vision. This also provides a simple and effective solution for adding meta-information using the transformer layer. 

Intuitively, species distribution presents a trend of clustering geographically, and the living habits of different species are different so that spatio-temporal information can assist the fine-grained task of species classification. When conditioned on latitude and longitude, we firstly want geographical coordinates to wrap around the earth. To achieve this, We converted the geographic coordinate system to a rectangular coordinate system, i.e., $[lat,lon] \to [x,y,z]$. Similarly, the distance between December and January is closer than the distance from October. And, 23:00 should result in a similar embedding with 00:00. Therefore, we perform the mapping $[month,hour] \to [sin(\frac{2 \pi month}{12}),cos(\frac{2 \pi month}{12}),sin(\frac{2 \pi hour}{24}),cos(\frac{2 \pi hour}{24})]$.  

When using attribute as meta-information, we initialize the attribute list as a vector. For example, there are 312 attributes on the CUB-200-2011 dataset; thus, a vector with a dimension of 312 can be generated. For meta-information in text form, we obtain the embedding of each word by BERT \cite{devlin2018bert}. In particular, when each image has multiple sentences as meta-information, we randomly select one sentence for training each time, and the maximum length of each sentence is 32. 

Further, as shown in Fig \ref{fig:overview}, non-linear embedding ($f:R^n \to R^d$) is a multi-layered fully-connected neural network that maps meta-information to embedding vector. Vision information and meta-information are different semantic levels. Thus, it is more difficult to learn visual information than auxiliary information. If a large amount of auxiliary information is fed to the network in the early stage of training, the visual ability of the network will be impaired. We mask part of the meta-information in a linearly decreasing ratio during the training to alleviate this problem.


\section{Experiments}
\textbf{Datasets.} We conduct experiments on ImageNet \cite{deng2009imagenet} image classification while it provides pre-trained models for fine-grained classification. We verify the effectiveness of our framework for adding meta-information on iNaturalist 2017 \cite{van2017inaturalist}, iNaturalist 2018 \cite{van2018inaturalist}, iNaturalist 2021 \cite{inat21}, and CUB-200-2011 \cite{wah2011caltech}. We also evaluate our proposed framework on several widely used fine-grained benchmarks, i.e., Stanford Cars \cite{krause20133d}, Aircraft \cite{maji2013fine}, and NABirds \cite{van2015building}. In addition, we do not use any bounding box/part annotation. The details of benchmarks widely used for fine-grained classification are summarized in Table \ref{table:datasets}.
\begin{table}[htbp]
\caption{Dataset statistics. Meta represents whether there is auxiliary information that can be used to improve the accuracy of fine-grained recognition.}
\label{table:datasets}
\setlength{\tabcolsep}{1.3mm}{
\begin{tabular}{c|c|c|c|c}
\hline
\textbf{Datasets}         & \textbf{Category} & \textbf{Meta}         & \textbf{Training}  & \textbf{Testing} \\ \hline
\textbf{iNaturalist 2017} & 5,089    & $\checkmark$ & 579,184   & 95,986  \\
\textbf{iNaturalist 2018} & 8,142    & $\checkmark$ & 437,513   & 24,426  \\
\textbf{iNaturalist 2021} & 10,000   & $\checkmark$ & 2,686,843 & 100,000 \\
\textbf{CUB-200-2011}     & 200      & $\checkmark$ & 5,994     & 5,794   \\
\textbf{Stanford Cars}    & 196      & \textbf{$\times$}     & 8,144     & 8,041   \\
\textbf{Aircraft}         & 100      & \textbf{$\times$}     & 6,667     & 3,333   \\
\textbf{NABirds}          & 555      & \textbf{$\times$}     & 23,929    & 24,633  \\ \hline
\end{tabular}}
\end{table}

\textbf{Implementation details.} First, we resize input images to 384*384. AdamW \cite{kingma2014adam} optimizer is employed with using a cosine decay learning rate scheduler. The learning rate is initialized as $5e^{-5}$ except $5e^{-3}$ for the Stanford Cars dataset and $5e^{-4}$ for the Aircraft dataset. The weight decay is 0.05. We include most of the augmentation and regularization strategies of \cite{liu2021swin} in training. We fine-tune the model for 300 epochs and perform 5 epochs of warm-up. An increasing degree of stochastic depth augmentation is employed for MetaFormer-0, MetaFormer-1, MetaFormer-2 with the maximum rate of 0.1, 0.2, 0.3, respectively.

\subsection{Comparison with CoAtNet on ImageNet-1k} 

\begin{table}[htbp]
\caption{Comparison with CoAtNet on ImageNet-1k. The results show that MetaFormer outperforms CoAtNet on ImageNet-1k. More comparisons with other SotA backbones could be found in appendix.}
\label{table:result in imagenet1k}
\setlength{\tabcolsep}{0.8mm}{
\begin{tabular}{c|ccc|c}
\hline
Method       & image size & \#Param. & \#FLOPS & \begin{tabular}[c]{@{}c@{}}ImageNet\\  top-1 acc\end{tabular} \\ \hline
CoAtNet-0 \cite{dai2021coatnet}    & $224^2$        & 25M      & 4.2G  & 81.6                                                          \\
CoAtNet-1 \cite{dai2021coatnet}    & $224^2$        & 42M      & 8.G   & 83.3                                                          \\
CoAtNet-2 \cite{dai2021coatnet}    & $224^2$        & 75M      & 15.7G & 84.1                                                          \\ \hline
MetaFormer-0 & $224^2$        & 28M      & 4.6G  & 82.9                                                          \\
MetaFormer-1 & $224^2$        & 45M      & 8.5G  & 83.9                                                          \\
MetaFormer-2 & $224^2$        & 81M      & 16.9G & 84.1                                                          \\ \hline

\end{tabular}}
\end{table}
Table \ref{table:result in imagenet1k} shows the accuracy of ImageNet-1k. Our network architecture outperforms CoAtNet. When we implement this architecture, CoAtNet is our template. The obvious difference between MetaFormer and CoAtNet is that MetaFormer retains the class token in the ViT to obtain the final output, while CoAtNet uses pooling. Especially, we additionally designed an aggregate layer to integrate class tokens obtained at different stages. Finally, Using regular ImageNet-1k training, MetaFormer can achieve performance that exceeds CoAtNet: $+2.3\%$ for MetaFormer-0 over CoAtNet-0, and $+0.6\%$ for MetaFormer-1 over CoAtNet-1, respectively.

\subsection{The Power of Meta Information}

\begin{table*}[htbp]
\caption{Results in iNaturalist 2019, iNaturalist 2018, and iNaturalist 2021 with meta-information. The green numbers represent the improvement brought by adding meta-information compared to only using images as input. It is worth noting that with the improvement of visual ability, the improvement brought by meta-information has not been greatly attenuated, which demonstrates the necessity of meta-information.}
\label{table:res in inat}
\setlength{\tabcolsep}{2.1mm}{
\begin{tabular}{c|cccc|lll}
\hline
Method                         & Backbone                      & Pre-training                  & \begin{tabular}[c]{@{}c@{}}Image\\ size\end{tabular} & Meta method  & iNat17     & iNat18     & iNat21     \\ \hline
\multirow{4}{*}{Geo-Aware \cite{chu2019geo}}     & \multirow{4}{*}{Inception V3} & \multirow{4}{*}{ImageNet-1k}  & \multirow{4}{*}{299}                                 & Image-Only   & 70.1       & -          & -          \\
                               &                               &                               &                                                      & Whitelisting & 72.6       & -          & -          \\
                               &                               &                               &                                                      & Post-Process & 79.0       & -          & -          \\
                               &                               &                               &                                                      & Feature Mod  & 78.2       & -          & -          \\ \hline
\multirow{4}{*}{Presence-Only \cite{mac2019presence}} & \multirow{4}{*}{Inception V3} & \multirow{4}{*}{ImageNet-1k}  & \multirow{2}{*}{299}                                 & Image-Only   & 63.27      & 60.2       & -          \\
                               &                               &                               &                                                      & Prior        & 69.6       & 72.7       & -          \\
                               &                               &                               & \multirow{2}{*}{520}                                 & Image-Only   & -          & 66.2       & -          \\
                               &                               &                               &                                                      & Prior        & -          & 77.5       & -          \\ \hline
\multirow{8}{*}{MetaFormer}        & \multirow{2}{*}{MetaFormer-0}    & \multirow{2}{*}{ImageNet-1k}  & \multirow{2}{*}{384}                                 & Image-Only   & 75.7       & 79.5       & 88.4       \\
                               &                               &                               &                                                      & Transformer  & 79.8\textcolor{mygreen}{ (+4.1)} & 85.4\textcolor{mygreen}{ (+5.9)} & 92.6\textcolor{mygreen}{ (+4.2)} \\ \cline{2-8} 
                               & \multirow{2}{*}{MetaFormer-1}    & \multirow{2}{*}{ImageNet-1k}  & \multirow{2}{*}{384}                                 & Image-Only   & 78.2       & 81.9       & 90.2       \\
                               &                               &                               &                                                      & Transformer  & 81.3\textcolor{mygreen}{ (+3.1)} & 86.5\textcolor{mygreen}{ (+4.6)} & 93.4\textcolor{mygreen}{ (+3.2)} \\ \cline{2-8} 
                               & \multirow{4}{*}{MetaFormer-2}    & \multirow{2}{*}{ImageNet-1k}  & \multirow{2}{*}{384}                                 & Image-Only   & 79.0       & 82.6       & 89.8       \\
                               &                               &                               &                                                      & Transformer  & 82.0\textcolor{mygreen}{ (+3.0)} & 86.8\textcolor{mygreen}{ (+4.2)} & 93.2\textcolor{mygreen}{ (+3.4)} \\ \cline{3-8} 
                               &                               & \multirow{2}{*}{ImageNet-21k} & \multirow{2}{*}{384}                                 & Image-Only   & 80.4       & 84.3       & 90.3       \\
                               &                               &                               &                                                      & Transformer  & 83.4\textcolor{mygreen}{ (+3.0)} & 88.7\textcolor{mygreen}{ (+4.4)} & 93.6\textcolor{mygreen}{ (+3.3)} \\ \hline
\end{tabular}}
\end{table*}
The table \ref{table:res in inat} shows the results of a series of iNaturalist datasets with spatio-temporal prior. Geo-Aware \cite{chu2019geo} systematically examined various ways of incorporating geolocation information into fine-grained image classification, such as whitelisting, post-processing, and feature modulation. Presence-Only \cite{mac2019presence} use spatio-temporal information as the prior to improve the accuracy of fine-grained recognition. Limited by the network architecture, previous advances on geographical priors were only carried out on the poor baseline. 

In this paper, we provide a series of strong baselines with spatio-temporal information. Moreover, we employ the transformer layer in the backbone to utilize additional information without any special head. In the case of different input sizes and different model sizes, adding spatio-temporal information in our way can achieve a consistent improvement of \textbf{3\%-6\%}. On the one hand, it shows the power of meta-information, and on the other hand, it shows the rationality of the way that MetaFormer adds meta-information. 

Moreover, when a larger model is used, the visual ability can be improved reasonably. For example, compared to MetaFormer-0, MetaFormer-2 increases the accuracy of the iNaturalist 2017 from $75.7\%$ to $79.0\%$ with model pre-trained on ImageNet-1k. A stronger pre-training model can also bring performance improvements. For example, when adopting MetaFormer-2, the accuracy of iNaturalist 2017 can be increased from $79.0\%$ to $80.4\%$ using a model pre-trained on ImageNet-21k. We have observed that the visual ability is improved while the gain brought by meta-information has not been greatly attenuated when using a larger model and stronger pre-training. This shows that part of the samples in the test set must be effectively identified with the aid of meta-information. In addition, MetaFormer achieved \textbf{83.4\%}, \textbf{88.7\%} and \textbf{93.6\%} accuracy on iNaturalist 2017, iNaturalist 2018 and iNaturalist 2021, respectively. This provides benchmark results for the iNaturalist series of large-scale datasets.

\begin{table}[htbp]
\caption{Result on CUB-200-2011 with meta-information. Image-Only represents using image only as input in training. Image+Attribute and Image+Text represent adding attribute and text description on the basis of the image as input in training. Input in Testing represents the format of the input information used in the testing. We observe that the addition of meta-information can not only improve the final performance of fine-grained recognition, but also improve the visual ability of the model on the CUB-200-2011.}
\label{table:res in CUB with text}
\setlength{\tabcolsep}{0.3mm}{
\begin{tabular}{c|c|cl}
\hline
Method                           & Backbone                  & \begin{tabular}[c]{@{}c@{}}Input \\ in Testing\end{tabular} & \ CUB        \\ \hline
ResNet-50 \cite{he2016deep}                        & ResNet-50                 & image                                                       & \ 84.5       \\
CVL \cite{he2017fine}                              & VGG-16                    & image+text                                                  & \ 85.6       \\
KERL \cite{chen2018knowledge}                             & VGG-16                    & image+attr                                             & \ 87.0       \\
S3N \cite{ding2019selective}                              & ResNet-50                 & image                                                       & \ 89.6       \\
StackedLSTM \cite{ge2019weakly}                      & GoogleNet                 & image                                                       & \ 90.4       \\
CAP \cite{behera2021context}                              & Xception                  & image                                                       & \ 91.8       \\ \hline
Image-Only                       & MetaFormer-1                  & image                                                       & \ 91.4       \\ \hline
\multirow{2}{*}{Image+Text} & \multirow{2}{*}{MetaFormer-1} & image                                                       & \ 91.7\textcolor{myred}{ (+0.3) } \\ 
                                 &                           & image+text                                             & \ 91.9\textcolor{mygreen}{ (+0.2) } \\ \hline
\multirow{2}{*}{Image+Attribute}      & \multirow{2}{*}{MetaFormer-1} & image                                                       & \ 91.5\textcolor{myred}{ (+0.1) } \\
                                 &                           & image+attr                                                  & \ 91.8\textcolor{mygreen}{ (+0.3) } \\ \hline
\end{tabular}}
\end{table}
In order to verify that our model can adapt to various forms of additional information, we conducted experiments on the CUB-200-2011 with text description as well as attribute. The results in the table \ref{table:res in CUB with text} show that the accuracy can be increased from 91.7\% to 91.9\% when using image and text description as input in testing. A similar result can be observed when using attributes as meta-information. To effectively ensure the validity of meta-information, we use a model pre-trained on Imagenet-21k to initialize the parameters of MetaFormer-1. In the case of a strong baseline, meta-information can still bring gain, which shows that our method indeed leverages meta-information to assist fine-grained recognition. 

CVL \cite{he2017fine} designed complex vision stream and language stream to leverage text descriptions to improve the accuracy of fine-grained recognition. KERL \cite{chen2018knowledge} integrates the knowledge graph into the feature learning to promote fine-grained image recognition, thereby using attribute information to supervise the learning. Compared with these methods that require complex modules, our method is straightforward and can adapt to different meta-information. Note that these methods are verified based on the poor baseline. In addition, CAP \cite{behera2021context} achieved the SotA performance on the CUB-200-2011. Our method can achieve comparable performance to CAP without meta-information. 

Using images as input in training, MetaFormer-1 achieves 91.4\% accuracy on CUB-200-2011. Under the same training settings, when image and text description are used as input in training and the only image is used as input in testing, the accuracy rate becomes 91.7\%. This shows that meta-information can not only improve the final recognition performance, but also promote the improvement of the model's visual ability.

\subsection{The Visualization of Meta Information}
\begin{figure*}[htbp]
    \begin{center}
        \includegraphics[scale=0.26]{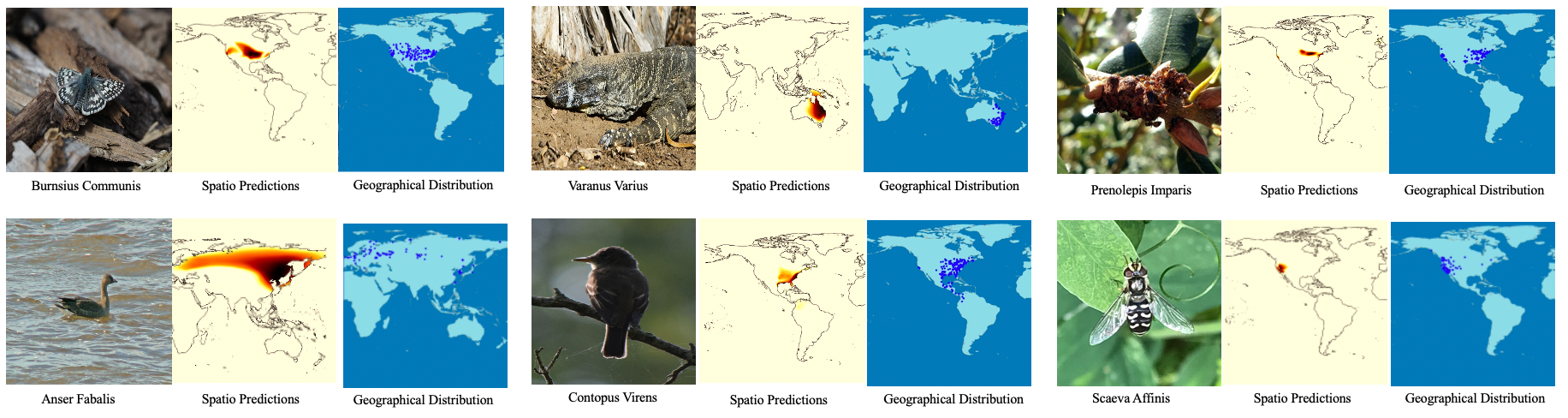}
    \end{center}
\caption{Spatial predictions. Predicted distributions for several object categories using a model trained on iNaturalist 2021. Darker color indicates that the current location is more responsive to the category of interest. Scattered points represent the true geographic distribution of the current species.}
\label{geography}
\end{figure*}

To have an intuitive understanding of meta-information, following \cite{mac2019presence}, we firstly generate spatial predictions for several different species from iNaturalist 2021. In Fig. \ref{geography}, each image is generated by querying each location on the surface of the earth to generate a prediction of the category of interest. The scattered points represent the true geographical distribution of the current species. In practice, we evaluate $1000\times2000$ spatial locations and mask out the predictions over the ocean for visualization. It can be seen from the visualization that the model can learn the geographic distribution of species and thus use the prior of this geographic distribution to assist fine-grained classification.

\begin{figure}[htbp]
    \begin{center}
        \includegraphics[scale=0.18]{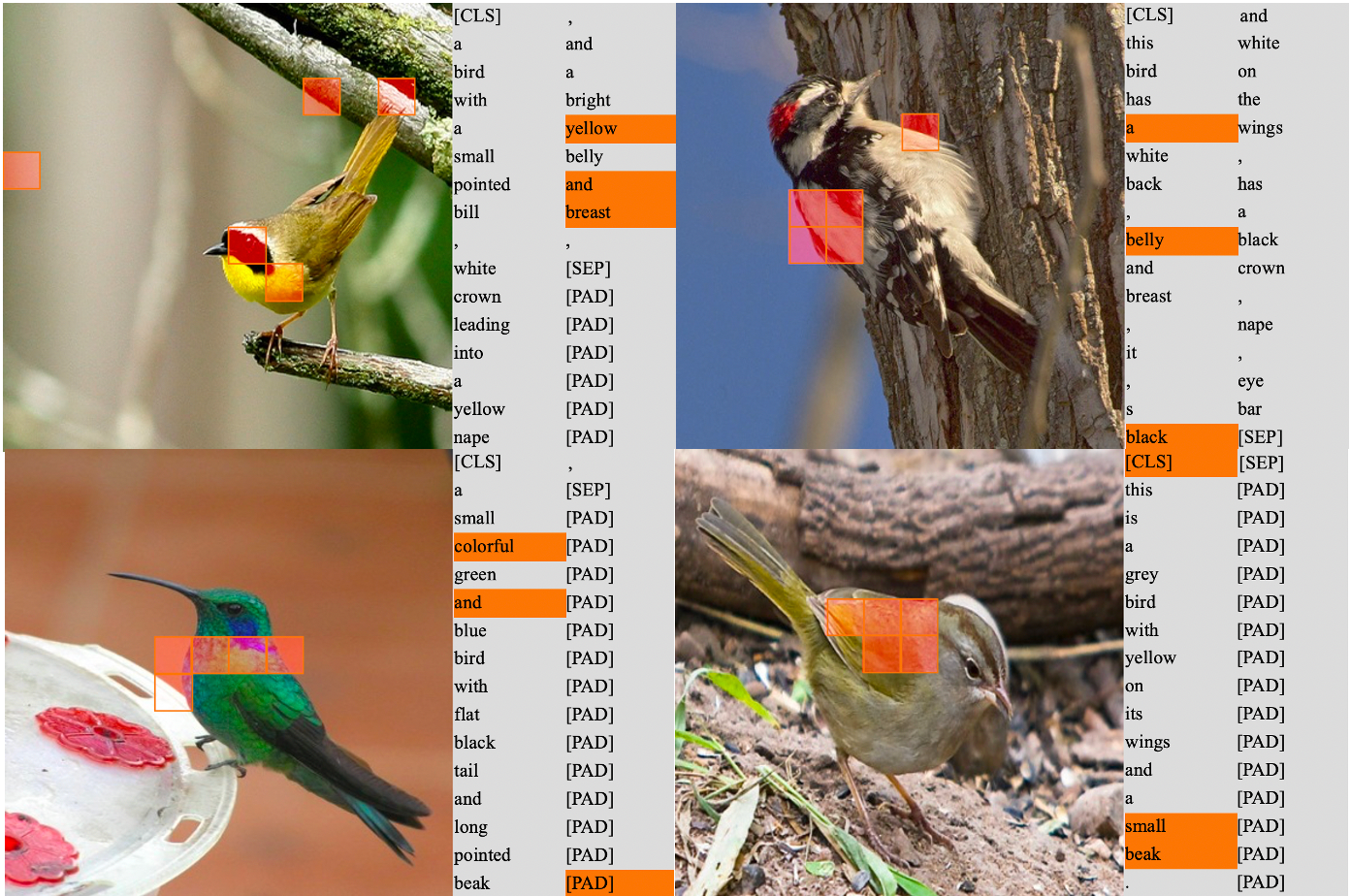}
    \end{center}
\caption{Top-k of similarity between class token with other tokens including vision token and word token. The orange squares in the image represent the five visual tokens that are most similar to the class token. In addition, the orange background in the text represents the three word tokens that are most similar to the class token.}
\label{language}
\end{figure}

\begin{figure}[htbp]
    \begin{center}
        \includegraphics[scale=0.28]{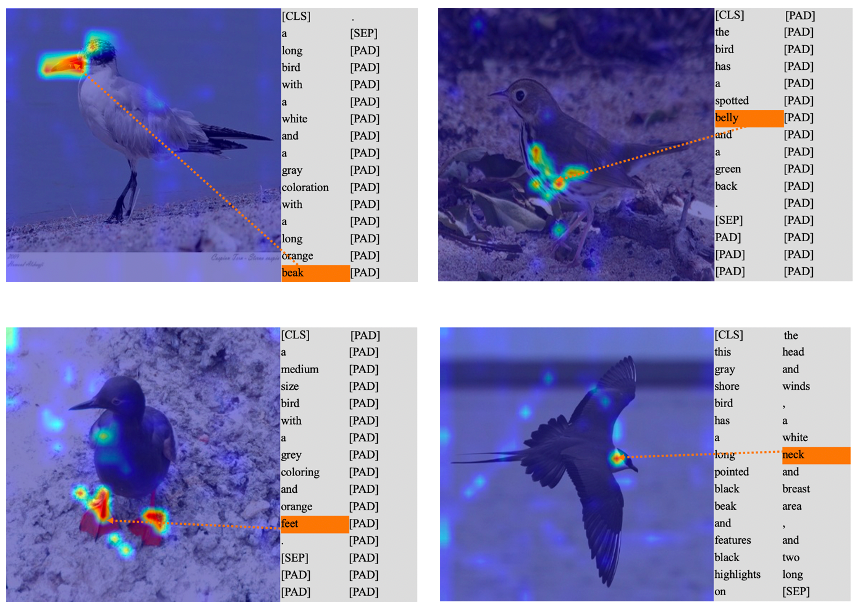}
    \end{center}
\caption{Self-attention map of word token. The warmer the color, the higher the similarity between the token of the current position and the word token.}
\label{heatmap}
\end{figure}

In order to verify whether the model uses the text information to assist fine-grained recognition, in Fig. \ref{language}, we visualize the top-5 of the similarity between the vision token and class token and the top-3 between the word token and class token, respectively. The class token is finally used to predict the species category. From the visualization, it can be seen that the class token has a high similarity with some tokens representing the species' attributes. Moreover, visual tokens and word tokens with high similarity often show a complementary relationship. Meanwhile, in Fig. \ref{heatmap}, we visualize the visual attention map corresponding to the word token, in which the words representing the attributes of the species usually have a high similarity with the corresponding vision token.

\subsection{The Importance of Pre-trained Model}
Pre-trained models are essential for fine-grained classification, but, to the best of our knowledge, no research has given a baseline for fine-grained classification under different pre-training. So in this paper, we study in detail the impact of varying pre-training on fine-grained classification and achieved SotA performance on several datasets.

The experiment results on CUB-200-2011 and NABirds are shown in Table \ref{table:res in cub and nabirds}. Compared to Imagenet-1k, when we transfer networks trained on Imagenet-21k,  MetaFormer-1 achieved 2.0\% and 2.2\% improvements on CUB-200-2011 and NABirds. The accuracy of CUB-200-2011 and NABirds is 92.3\% and 92.7\%,  respectively, which outperforms the SotA approaches (91.8\% and 91.0\% on CAP\cite{behera2021context}) by a clear margin, using iNaturalist 2021 for pre-training. iNaturalist 2021 with fewer data can perform better than Imagenet-21k since the domain similarity between iNaturalist 2021 and downstream datasets is higher. Using the MetaFormer-0 with fewer parameters and models pre-trained on iNaturalist, we also achieve performance (91.8\% and 91.2\%) equivalent to the SotA approaches. 

Existing methods are designed with complex multi-stage strategies (CPM \cite{ge2019weakly}), multi-branch structures (Cross-X \cite{luo2019cross}, API-Net \cite{zhuang2020learning}) or elaborate attention modules (CAL \cite{rao2021counterfactual}, CAP \cite{behera2021context}), resulting in difficulty in implementing. DSTL \cite{cui2018large}studies transfer learning by fine-tuning from large-scale datasets to small-scale datasets and carefully selects the data used for pre-training. Our experiments show that when the amount of data used for pre-training is higher and there are more categories, better performance can be achieved without selecting data. However, we did not deliberately select data during pre-training. FixSENet-154 \cite{touvron2019fixing} designed a complex image resolution strategy for training and testing, and we use a scientific image resolution strategy. When ImageNet-21k is also used to pre-train the model, our method achieves the same performance as TransFG \cite{he2021transfg} without any additional structure, and our model has fewer parameters and higher throughput. Our experimental results show that the SotA performance can still be achieved on the CUB-200-2011 and NABirds datasets without any inductive bias of fine-grained recognition tasks. This can provide researchers with a simple and effective baseline model and facilitate actual implementation.

\begin{table}[htbp]
\caption{Results on CUB-200-2011 and NABirds with different pre-trained models.}
\label{table:res in cub and nabirds}
\setlength{\tabcolsep}{0.3mm}{
\begin{tabular}{c|cc|cc}
\hline
Method                & Backbone                & Pretain      & CUB & NABirds \\ \hline
CPM \cite{ge2019weakly}                   & GoogleNet               & ImageNet-1k  & 90.4    & -       \\
CAL \cite{rao2021counterfactual}                   & ResNet101               & ImagNet-1k  & 90.6    & -       \\
TransFG \cite{he2021transfg}               & ViT-B\_16               & ImageNet-21k & 91.7    & 90.8    \\
CAP \cite{behera2021context}                   & Xception                & ImageNet-1k  & 91.8    & 91.0    \\
Cross-X \cite{luo2019cross}               & ResNet50                & ImageNet-1k  & 87.7    & 86.2    \\
DSTL \cite{cui2018large}                  & Inception-v3            & iNat17       & 89.3    & 87.9    \\
API-Net \cite{zhuang2020learning}               & DenseNet-161            & ImageNet-1k  & 90.0    & 88.1    \\
FixSENet \cite{touvron2019fixing}          & SENet-154               & ImageNet-1k  & 88.7    & 89.2    \\ \hline
\multirow{4}{*}{MetaFormer} & MetaFormer-0                  & iNat21         & 91.8    & 91.2   \\
                      & \multirow{3}{*}{MetaFormer-1} & ImageNet-1k  & 89.7    & 89.4    \\
                      &                         & ImageNet-21k & 91.3    & 91.6    \\
                      &                         & iNat21       & \textbf{92.3}    & \textbf{92.7}    \\ \hline
\end{tabular}}
\end{table}

\begin{table}[htbp]
\caption{Results on iNaturalist 2017 and iNaturalist 2018 with different pre-trained models.}
\label{table:res in inat17 and inat18}
\setlength{\tabcolsep}{0.3mm}{
\begin{tabular}{c|cc|cc}
\hline
Method                  & Backbone                  & Pretain      & iNat17 & iNat18 \\ \hline
TransFG \cite{he2021transfg}                 & ViT-B\_16                 & ImageNet-21k & 70.9   & -      \\
FixSENet \cite{touvron2019fixing}            & SENet-154                 & ImageNet-1k  & 75.4   & -      \\
DeiT-B \cite{touvron2021training}                  & ViT-B\_16                 & ImageNet-21k & -      & 80.1   \\
Grafit \cite{touvron2021grafit}                  & RegNet-8GF                & ImageNet-1k  & -      & 81.2   \\ \hline
\multirow{3}{*}{MetaFormer} & \multirow{3}{*}{MetaFormer-1} & ImageNet-1k  & 78.2   & 81.9   \\
                        &                           & ImageNet-21k & 79.4   & 83.2   \\
                        &                           & iNat21       & \textbf{82.0}   & \textbf{87.5}   \\ \hline
\end{tabular}}
\end{table}

iNaturalist 2017 and iNaturalist 2018 are large-scale datasets for fine-grained recognition. 
In Table \ref{table:res in inat17 and inat18}, we show the SotA results on iNaturalist 2017 and iNaturalist 2018, using MetaFormer-1. We observe that there is currently no reference performance for both iNaturalist 2017 and iNaturalist 2018. For example, when the model parameters trained by ImageNet-1k are used to initialize the model, FixSENet \cite{touvron2019fixing} achieves an accuracy of 75.4\% on iNaturalist 2017 and Grafit \cite{touvron2021grafit} achieves an accuracy of 81.2\% on iNaturalist 2018. However, our experiment found that the accuracy of iNaturalist 2017 and iNaturalist 2018 should be 78.2\% and 81.9\%, respectively, without any special design, using the model pre-trained on ImageNet-1k. The transfer learning performance by fine-tuning MetaFormer-1 on fine-grained datasets is also presented in Table \ref{table:res in inat17 and inat18}. More results can be found in the appendix.

\begin{table}[htbp]
\caption{Results on Stanford Cars and Aircraft with different pre-trained models.}
\label{table:res in cars and aircraft}
\setlength{\tabcolsep}{0.53mm}{
\begin{tabular}{c|cc|cc}
\hline
Method                  & Backbone                  & Pretain      & Cars & Aircraft \\ \hline
GPipe \cite{huang2019gpipe}                   & AmoebaNet-B               & ImageNet-1k  & 94.6 & 92.7     \\
DCL \cite{chen2019destruction}                     & ResNet-50                 & ImageNet-1k  & 94.5 & 93.0     \\
S3N \cite{ding2019selective}                     & ResNet-50                 & ImageNet-1k  & 94.7 & 92.8     \\
PMG \cite{du2020fine}                     & ResNet-50                 & ImageNet-1k  & 95.1 & 93.4     \\
API-Net \cite{zhuang2020learning}                 & DenseNet-161              & ImageNet-1k  & 95.3 & 93.9     \\
CAP \cite{behera2021context}                     & Xception                  & ImageNet-1k  & \textbf{95.7} & 94.1     \\ \hline
\multirow{3}{*}{MetaFormer} & \multirow{3}{*}{MetaFormer-1} & ImageNet-1k  & 94.9    & 92.8        \\
                        &                           & ImageNet-21k & 95.0 & 94.2        \\
                        &                           & iNat21       & 95.0    & \textbf{94.3}        \\ \hline
\end{tabular}}
\end{table}
Table \ref{table:res in cars and aircraft} shows the results of our model on Stanford Cars and Aircraft. On Stanford Cars and Aircraft, most of the previous methods used ImageNet-1k for pre-training. We offer the different transfer learning performances by fine-tuning MetaFormer-1 on these two fine-grained datasets. Experiments show that on Stanford Cars, a more potent pre-training model does not bring further performance improvement. We argue that more simple pictures in the Stanford Cars dataset require less work on pre-trained models. On the Aircraft dataset, the model pre-trained with iNaturalist 2021 is worse than that trained with ImageNet-21k because it has a more extensive domain gap with the downstream domain.


\section{Conclusion}
In this work, we propose a unified meta-framework for fine-grained visual classification. MetaFormer uses the transformer to fuse visual information and various meta-information, not introducing any additional structure. Meanwhile, MetaFormer also provides a simple yet effective baseline for FGVC. In addition, we systematically examined the impact of different pre-training models on fine-grained tasks. MetaFormer achieves SotA performance on the iNaturalist series, CUB-200-2011, and NABirds datasets. Meanwhile, we believe that meta-information is essential for fine-grained recognition tasks in the future. And, MetaFormer can provide a way to utilize various auxiliary information.


\appendix

\setcounter{page}{1}

\twocolumn[
\centering
\Large
\textbf{MetaFormer : A Unified Meta Framework for Fine-Grained Recognition} \\
\vspace{0.5em}Supplementary Material \\
\vspace{1.0em}
] 
\appendix
\section{The detailed information of MetaFormer}
\label{detail info of MetaFormer}
\textbf{Detailed experimental setting for ImageNet-1k and ImageNet-21k.} When training from the scratch on ImageNet-1k, the input image size is $224^2$. we adopt AdamW \cite{kingma2014adam} optimizer and train for 300 epochs and 20 epochs of linear warm-up with batchsize of 1024. The learning rate is initialized as $1e^{-3}$ and weight decay is 0.05. Most of the augmentation and regularization strategies of \cite{liu2021swin} are included in training. Note that an increasing degree of stochastic depth augmentation is employed for larger models, i.e. 0.1, 0.2, 0.3 for MetaFormer-0, MetaFormer-1, and MetaFormer-2, respectively. For resolutions of $384^2$, we fine-tune the models trained at $224^2$ resolution using an initial learning rate of $1e^{-4}$ for 30 epochs and 2 epochs of warm-up, instead of training from scratch. For ImageNet-21k, we train for 90 epochs and 5 epochs of warm-up with the input image resolution of $224^2$ and fine-tune a model for 10 epochs with the input image resolution of $384^2$.

\textbf{Detailed architecture of MetaFormer.} The MetaFormer consists of the convolutional layer and the transformer layer. The first three stages mainly adopt MBConv blocks, and the latter two stages adopt the Relative transformer blocks.  We mimic the canonical convolutional network, adpot the convolution layer with stride of 2 in stage 0 and stage 1 for downsampling, and adopt max-pooling for downsampling in stage 2. In stage 3 and stage 4, overlapping patch embedding is employed for downsampling. The class tokens of stage 3 and stage 4 are integrated through the aggregate layer. Among them, the class token of stage3 will be dimensionally expanded by MLP. For all Transformer blocks, the size of each attention head is 8. The expansion rate for the inverted bottleneck is always 4, and the expansion (shrink) rate for the Squeeze-and-Excitation is always 0.25.

\textbf{Performance comparison with SotA backbone.} Parameters, flops and throughput of MetaFormer are shown in the table \ref{table:res backbone}. Meanwhile, it shows the comparison result on ImageNet-1k with the state-of-the-art backbone. 

\textbf{Performance comparison of CLT and GAP.} We ultimately design a simple and effective framework, which can integrate a variety of meta information. Therefore, we retain the class token as a bridge between visual information and additional prior information. The class token can pass through S3 and S4 in serial ($CLT_{serial}$), or in parallel ($CLT_{parallel}$). Specifically, the parallel means that S3 and S4 obtain two class tokens, respectively, and then they are combined through the aggregate layer. The ablation study is shown in the table \ref{table:diff of cls}. In table \ref{table:diff of cls}, $GAP$ represents the global average pooling operation, and $CLT_{final}$ represents only the S4 class token is used for class prediction. Experiments show that the result of $CLT_{parallel}$  using a aggregate layer is better than $CLT_{final}$ and $CLT_{serial}$. Moreover, using GAP is not better than using class token on ImageNet-1k. 
\begin{table}[htbp]
\caption{Accuracy of MetaFormer using different methods for class prediction. GAP represents performing global average pooling to obtain the feature vector for classification prediction. CLT means leveraging the class token to classify.}
\label{table:diff of cls}
\setlength{\tabcolsep}{5mm}{
\begin{tabular}{l|c|c}
\hline
\multicolumn{1}{c|}{} & Backbone & \begin{tabular}[c]{@{}c@{}}ImageNet\\ top-1 acc\end{tabular} \\ \hline
$GAP$                   & MetaFormer-0   & 82.9                                                              \\ \hline
$CLT_{final}$                   & MetaFormer-0   & 82.6                                                            \\
$CLT_{serial}$                   & MetaFormer-0   & 82.8                                                            \\
$CLT_{parallel}$                   & MetaFormer-0   & 82.9                                                            \\ \hline
\end{tabular}}
\end{table}

\begin{figure*}[htbp]
    \begin{center}
        \includegraphics[scale=0.24]{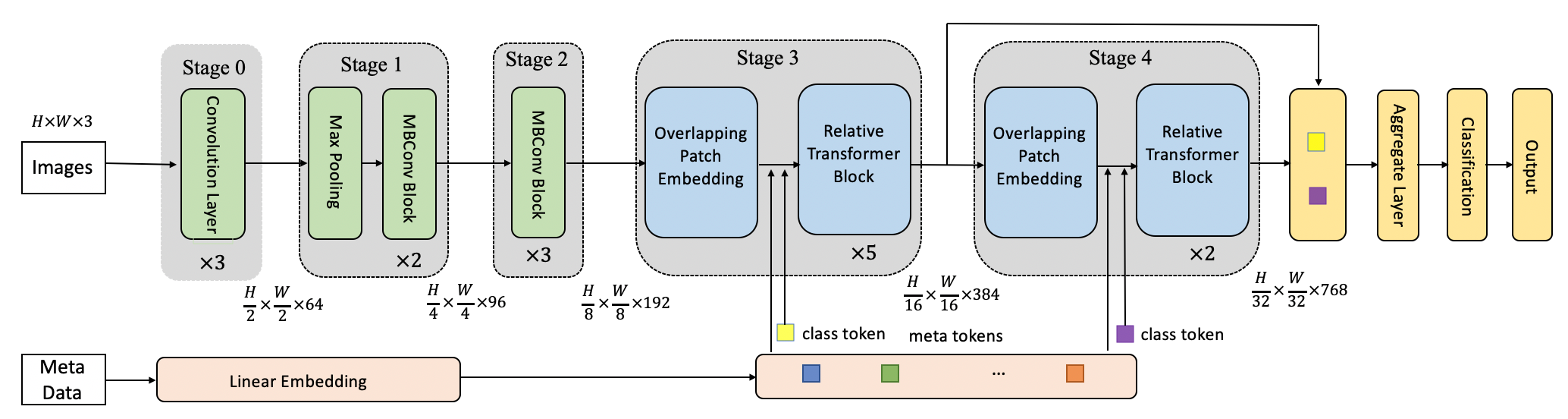}
    \end{center}
\caption{Overview of MetaFormer. The first three stages use convolution to downsample, and the next two stages use a relative transformer layer to fuse the image and meta information. The class tokens obtained in the two stages are fused through the aggregation layer.}
\label{fig:stucture of meta}
\end{figure*}

\begin{table*}[htbp]
\caption{The result of MetaFormer and comparison of other backbones on ImageNet-1k. Throughput is measured using the GitHub repository of \cite{rw2019timm} with V100 GPU}
\label{table:res backbone}
\setlength{\tabcolsep}{4.5mm}{
\begin{tabular}{lc|cccc|c}
\hline
\multicolumn{2}{c|}{Method}                   & \begin{tabular}[c]{@{}c@{}}Image \\ size\end{tabular} & \#Param. & \#FLOPS & \begin{tabular}[c]{@{}c@{}}Throughput\\  (image/s)\end{tabular} & \begin{tabular}[c]{@{}c@{}}ImageNet\\  top-1 acc\end{tabular} \\ \hline
\multirow{6}{*}{Conv only} & EfficientNet-B4 \cite{tan2019efficientnet}  & $380^2$                                                   & 19M      & 4.2G  & 349.4                                                           & 82.9                                                          \\
                           & EfficientNet-B5 \cite{tan2019efficientnet}  & $456^2$                                                   & 30M      & 9.9G  & 169.1                                                           & 83.6                                                          \\
                           & EfficientNet-B6 \cite{tan2019efficientnet}  & $528^2$                                                   & 43M      & 19.0G & 96.9                                                            & 84.0                                                          \\
                           & EfficientNet-B7 \cite{tan2019efficientnet}  & $600^2$                                                   & 66M      & 37.0G & 55.1                                                            & 84.3                                                          \\
                           & EfficientNetV2-S \cite{tan2021efficientnetv2} & $128^2-300^2$                                               & 24M      & 8.8G  & 666.7                                                           & 83.9                                                          \\
                           & EfficientNetV2-M \cite{tan2021efficientnetv2} & $128^2-380^2$                                               & 55M      & 24G   & 280.7                                                           & 85.1                                                          \\ \hline
\multirow{4}{*}{ViT only}  & ViT-B/16 \cite{dosovitskiy2020image}         & $384^2$                                                   & 86M      & 55.4G & 85.9                                                            & 77.9                                                          \\
                           & DeiT-S \cite{touvron2021training}           & $224^2$                                                   & 22M      & 4.6G  & 940.4                                                           & 79.8                                                          \\
                           & DeiT-B \cite{touvron2021training}           & $224^2$                                                   & 86M      & 17.5G & 292.3                                                           & 81.8                                                          \\
                           & DeiT-B \cite{touvron2021training}           & $384^2$                                                   & 86M      & 55.4G & 85.9                                                            & 83.1                                                          \\ \hline
\multirow{3}{*}{Local MSA} & Swin-T \cite{liu2021swin}           & $224^2$                                                   & 29M      & 4.5G  & 755.2                                                           & 81.3                                                          \\
                           & Swin-S \cite{liu2021swin}           & $224^2$                                                   & 50M      & 8.7G  & 436.9                                                           & 83.0                                                          \\
                           & Swin-B \cite{liu2021swin}           & $224^2$                                                   & 88M      & 15.4G & 278.1                                                           & 83.3                                                          \\ \hline
\multirow{6}{*}{Conv+MSA}  & CoAtNet-0 \cite{dai2021coatnet}        & $224^2$                                                   & 25M      & 4.2G  & -                                                               & 81.6                                                          \\
                           & CoAtNet-1 \cite{dai2021coatnet}        & $224^2$                                                   & 42M      & 8.4G  & -                                                               & 83.3                                                          \\
                           & CoAtNet-2 \cite{dai2021coatnet}        & $224^2$                                                   & 75M      & 15.7G & -                                                               & 84.1                                                          \\
                           & CoAtNet-0 \cite{dai2021coatnet}        & $384^2$                                                   & 25M      & 13.4G & -                                                               & 83.9                                                          \\
                           & CoAtNet-1 \cite{dai2021coatnet}        & $384^2$                                                   & 42M      & 27.4G & -                                                               & 85.1                                                          \\
                           & CoAtNet-2 \cite{dai2021coatnet}        & $384^2$                                                   & 75M      & 49.8G & -                                                               & 85.7                                                          \\ \hline
\multirow{6}{*}{Conv+MSA}  & MetaFormer-0         & $224^2$                                                   & 28M      & 4.6G  & 840.1                                                           & 82.9                                                          \\
                           & MetaFormer-1         & $224^2$                                                   & 45M      & 8.5G  & 444.8                                                           & 83.9                                                          \\
                           & MetaFormer-2         & $224^2$                                                   & 81M      & 16.9G & 438.9                                                           & 84.1                                                          \\
                           & MetaFormer-0         & $384^2$                                                   & 28M      & 13.4G & 349.4                                                           & 84.2                                                          \\
                           & MetaFormer-1         & $384^2$                                                   & 45M      & 24.7G & 165.3                                                           & 84.4                                                          \\
                           & MetaFormer-2         & $384^2$                                                   & 81M      & 49.7G & 132.7                                                           & 84.6                                                          \\ \hline
\end{tabular}}
\end{table*}

\section{Performance on fine-grained datasets with different pre-trained model. Large-scale pre-training can effectively improve the performance of fine-grained recognition.}
\label{diff pre-trained model}
\begin{table*}[htbp]
\caption{Result on fine-grained datasets with different pre-trained model}
\label{table:res with diff pretrain}
\setlength{\tabcolsep}{3.5mm}{
\begin{tabular}{c|c|cccccc}
\hline
Backbone                  & Pretrain     & CUB  & NABirds & iNaturalist 2017 & iNaturalist 2018 & Cars & Aircraft \\ \hline
\multirow{3}{*}{MetaFormer-0} & ImageNet-1k  & 89.6 & 89.1    & 75.7   & 79.5   & 95.0    & 91.2        \\
                          & ImageNet-21k & 89.7 & 89.5    & 75.8   & 79.9   & 94.6    & 91.2        \\
                          & iNaturalist 2021      & 91.8 & 91.5    & 78.3   & 82.9   & 95.1    & 87.4        \\ \hline
\multirow{3}{*}{MetaFormer-1} & ImageNet-1k  & 89.7 & 89.4    & 78.2   & 81.9   & 94.9    & 90.8        \\
                          & ImageNet-21k & 91.3 & 91.6    & 79.4   & 83.2   & 95.0    & 92.6        \\
                          & iNaturalist 2021      & 92.3 & 92.7    & 82.0   & 87.5   & 95.0    & 92.5        \\ \hline
\multirow{3}{*}{MetaFormer-2} & ImageNet-1k  & 89.7 & 89.7    & 79.0   & 82.6   & 95.0    & 92.4        \\
                          & ImageNet-21k & 91.8 & 92.2    & 80.4   & 84.3   & 95.1    & 92.9        \\
                          & iNaturalist 2021      & 92.9 & 93.0    & 82.8   & 87.7   & 95.4    & 92.8        \\ \hline
\end{tabular}}
\end{table*}
The table \ref{table:res with diff pretrain} shows the transfer performance of 6 fine-grained datasets(CUB-200-2011, NABirds, iNaturalist 2017, iNaturalist 2018, Stanford Cars, and Aircraft) under different pre-trained models. 


{
    \small
    \bibliographystyle{ieee_fullname}
    \bibliography{macros,main}
}


\end{document}